\documentclass[10pt,twocolumn,letterpaper]{article}

\usepackage{cvpr}              % To produce the CAMERA-READY version
\usepackage{graphicx}
\usepackage{amsmath}
\usepackage{amssymb}
\usepackage{booktabs}
\usepackage[shortlabels]{enumitem}
\usepackage[accsupp]{axessibility}  % Improves PDF readability for those with disabilities.

\usepackage[pagebackref,breaklinks,colorlinks]{hyperref}

\usepackage[capitalize]{cleveref}
\crefname{section}{Sec.}{Secs.}
\Crefname{section}{Section}{Sections}
\Crefname{table}{Table}{Tables}
\crefname{table}{Tab.}{Tabs.}

\def\confName{CVPR}
\def\confYear{2026}

\begin{document}

\title{EffiMiniVLM: A Compact Dual-Encoder Regression Framework}

\author{
\begin{tabular}{ccc}
Yin-Loon Khor$^{*}$ & Yi-Jie Wong$^{*\dagger}$ & Yan Chai Hum$^{\dagger}$ \\
Universiti Malaya & Universiti Tunku Abdul Rahman & Universiti Tunku Abdul Rahman \\
Kuala Lumpur 50603, Malaysia & Selangor 43000, Malaysia & Selangor 43000, Malaysia \\
{\tt\small yinloonkhor@gmail.com} & {\tt\small yjwong1999@gmail.com} & {\tt\small humyc@utar.edu.my}
\end{tabular}\\[0.5em]
{\small $^{*}$ indicates co-first author, $^{\dagger}$ indicates corresponding author}
}
\maketitle

\begin{abstract}
Predicting product quality from multimodal item information is critical in cold-start scenarios, where user interaction history is unavailable and predictions must rely on images and textual metadata. However, existing vision–language models typically depend on large architectures and/or extensive external datasets, resulting in high computational cost. To address this, we propose EffiMiniVLM, a compact dual-encoder vision–language regression framework that integrates an EfficientNet-B0 image encoder and a MiniLM-based text encoder with a lightweight regression head. To improve training sample efficiency, we introduce a weighted Huber loss that leverages rating counts to emphasize more reliable samples, yielding consistent performance gains. Trained using only 20\% of the Amazon Reviews 2023 dataset, the proposed model contains 27.7M parameters and requires 6.8 GFLOPs, yet achieves a CES score of 0.40 with the lowest resource cost in the benchmark. Despite its small size, it remains competitive with significantly larger models, achieving comparable performance while being approximately 4× to 8× more resource-efficient than other top-5 methods and being the only approach that does not use external datasets. Further analysis shows that scaling the data to 40\% alone allows our model to overtake other methods, which use larger models and datasets, highlighting strong scalability despite the model’s compact design.

\end{abstract}


\section{Introduction}

Predicting product quality from multimodal item information is closely related to the cold-start recommendation setting, where reliable user-side interaction history may be limited or unavailable. In such scenarios, product quality must be inferred primarily from item-side signals, such as images and textual metadata, rather than collaborative behavioural patterns. This makes the task both practically important and technically challenging, since visual appearance and textual descriptions often provide complementary but noisy evidence about product quality and user perception. Recent multimodal learning approaches have demonstrated the effectiveness of combining vision and language representations for downstream prediction tasks \cite{radford2021learningtransferablevisualmodels, yao2021filipfinegrainedinteractivelanguageimage, lu2019vilbertpretrainingtaskagnosticvisiolinguistic, chen2020uniteruniversalimagetextrepresentation, li2023blip2bootstrappinglanguageimagepretraining, chen2023palijointlyscaledmultilinguallanguageimage}. However, strong performance is often associated with increasingly large backbones and high computational cost \cite{johnson2017billionscalesimilaritysearchgpus}. In practical settings, especially under competition or deployment constraints, predictive accuracy alone is insufficient and resource efficiency is also a critical consideration. Therefore, an effective solution should achieve a favourable trade-off between predictive quality and computational cost. Despite this need, relatively limited work has focused on lightweight multimodal regression models tailored to item-only product quality prediction. This gap motivates the development of compact architectures that can operate effectively under strict resource constraints.

To address this challenge, we develop EffiMiniVLM, a compact dual-encoder regression framework that combines an EfficientNet-B0 image encoder with a MiniLM-based text encoder to jointly model product images and metadata. Instead of relying on large-scale architectures or external training data, our design focuses on maintaining a lightweight and efficient pipeline while still capturing useful multimodal information for rating prediction. The experimental results show that this resource-efficient design remains competitive and further analysis suggests that the framework retains substantial headroom for improvement as the training data scale increases. This work is submitted as our solution to the CVPR 2026 Challenge on Efficient VLM for Multimodal Creative Quality Scoring.\footnote{Code available at \url{https://github.com/yinloonkhor/CVPR2026-EffiMiniVLM}.}

\section{Related Works}

\subsection{Multimodal Learning}
Prior work in multimodal learning explores a spectrum of fusion strategies, ranging from coordinated (late fusion) to fully joint architectures. Early approaches often adopt late fusion, where modality-specific encoders are trained independently and combined at the representation or decision level. This paradigm is exemplified by dual-encoder models such as CLIP \cite{radford2021learningtransferablevisualmodels}, which learn aligned embeddings via contrastive objectives. Follow-up works, including FILIP \cite{yao2021filipfinegrainedinteractivelanguageimage}, extend this design with finer-grained matching while maintaining encoder independence. In contrast, intermediate and early fusion approaches introduce explicit cross-modal interaction, as in cross-attention architectures (e.g., ViLBERT \cite{lu2019vilbertpretrainingtaskagnosticvisiolinguistic}) and unified transformers (e.g., UNITER \cite{chen2020uniteruniversalimagetextrepresentation}), enabling richer token-level alignment. More recent vision–language models further integrate generative objectives and large language models, as seen in approaches such as BLIP-2 \cite{li2023blip2bootstrappinglanguageimagepretraining} and PaLI \cite{chen2023palijointlyscaledmultilinguallanguageimage}, which couple pretrained vision encoders with powerful language decoders to support multimodal reasoning and instruction following. Despite its simplicity, late fusion remains widely adopted due to its scalability and efficiency: independent encoding supports large-scale training and fast retrieval via nearest-neighbour search \cite{johnson2017billionscalesimilaritysearchgpus, radford2021learningtransferablevisualmodels}, facilitates modular design and transfer \cite{zhai2022litzeroshottransferlockedimage, pham2023combinedscalingzeroshottransfer}, and avoids the quadratic cost of cross-modal attention. These properties make late fusion a dominant paradigm in modern multimodal learning and a key foundation for recent vision–language models.

\subsection{Product Rating Prediction}
Product rating prediction has been widely studied using both product- and review-side information. Early methods rely on contextual signals, including user history and review context, to improve prediction accuracy. For instance, Amirifar et al. \cite{10183868} show that such context-aware approaches benefit from user- and review-level information, but are less suitable for strict cold-start scenarios where these signals are unavailable at inference time. They further propose a three-stage framework that combines general product attributes with customer-salient features extracted from reviews via named entity recognition, demonstrating that enhanced product-side features improve rating estimation \cite{10183868}. More recently, multimodal learning has emerged as a promising direction for capturing complementary signals from different data sources. Su et al. \cite{Su_2023} show that combining textual, visual, and structured inputs yields consistent improvements over unimodal models in vehicle rating prediction, while Han et al. \cite{han2022sanclmultimodalreviewhelpfulness} demonstrate in a related regression task that integrating text and image features via attention and contrastive learning improves prediction quality with efficient representations. Taken together, these findings suggest that multimodal item representations are particularly advantageous in cold-start settings where collaborative user signals are absent. However, relatively limited work has focused on lightweight multimodal regression models specifically tailored for item-only product rating prediction.

\section{Methodology}

\subsection{Model Architecture}

Unlike other teams, our proposed model, EffiMiniVLM, is designed with efficiency as the primary objective. The model has the lowest resource cost of 0.1, containing only 27.7M parameters and requiring merely 6.8 GFLOPs. Despite this extremely small computational footprint, EffiMiniVLM achieves a final CES score of 0.4 and is the only model within the top five positions with a resource cost below 0.4. This result demonstrates that carefully designed lightweight architectures can remain competitive with significantly larger models.

\begin{figure*}[t]
    \centering
    \includegraphics[width=\textwidth]{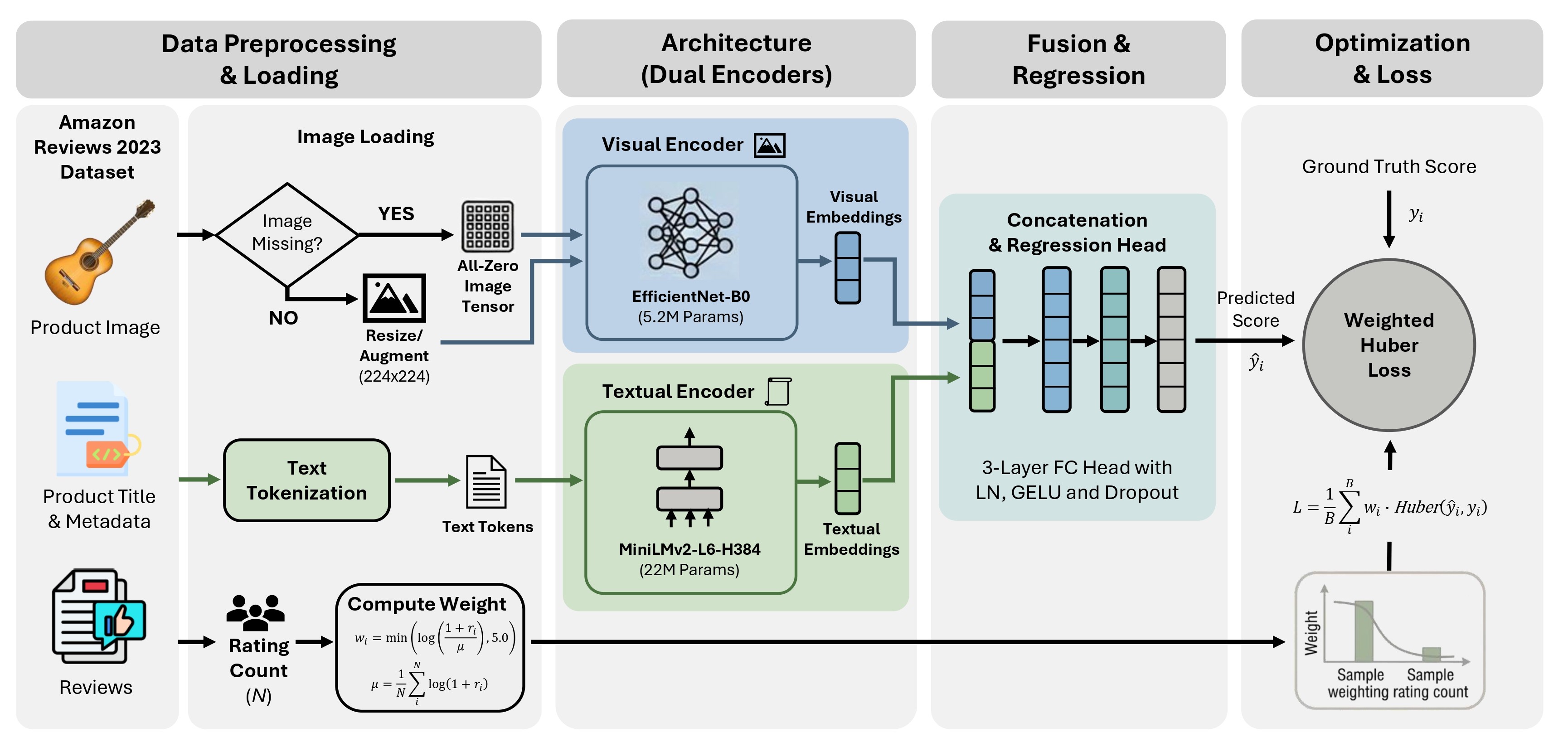}
    \caption{Model architecture of the proposed EffiMiniVLM.}
    \label{fig:model_architecture}
\end{figure*}

EffiMiniVLM follows a simple yet effective design. It is a dual-encoder vision–language regression model that predicts product quality scores from both image and text inputs. For the visual branch, EffiMiniVLM employs a pretrained EfficientNet-B0 \cite{tan2020efficientnetrethinkingmodelscaling} to extract compact and discriminative image features. This backbone is selected for its strong balance between representational capacity and computational efficiency. For the textual branch, a pretrained MiniLMv2-L6-H384 transformer encoder \cite{wang2021minilmv2multiheadselfattentionrelation} processes product metadata to produce dense semantic embeddings that capture contextual and descriptive information not directly observable from images. The visual and textual embeddings are concatenated to form a joint multimodal representation, which is then passed to a lightweight multilayer perceptron regression head to predict a scalar product quality score. No ensembling or advanced multimodal fusion module is used. By leveraging the complementary nature of visual and textual information, the model integrates visual cues from product images with semantic information extracted from structured textual metadata within a unified multimodal framework. An overview of the model is shown in Figure \ref{fig:model_architecture}.

To improve robustness in real-world scenarios, EffiMiniVLM is designed to handle incomplete multimodal inputs. When image data are unavailable, the model can still operate using only the textual branch, enabling reliable predictions under missing-modality conditions. Overall, EffiMiniVLM integrates an efficient convolutional image encoder, a compact transformer-based text encoder and a lightweight fusion-based regression head into a flexible and resource-efficient multimodal prediction framework. 

\subsection{Dataset}

For this challenge, we utilize only the Amazon Reviews 2023 dataset for training \cite{hou2024bridginglanguageitemsretrieval} without any additional data. We select 25 categories and preprocess them carefully in accordance with the challenge’s input constraints. Each training instance is constructed by concatenating the main category, title, features and description of an item. To improve data quality, we keep only samples with complete essential metadata, including the main category, title, features, description, average rating and rating number. We adopt a preprocessing strategy similar to that in \cite{hou2024bridginglanguageitemsretrieval}, which includes removing special characters such as tabs, newlines and carriage returns, as well as excluding instances with contextual or item metadata shorter than 30 characters. Due to computational constraints, only 20\% of the full training set is used. Following evidence that balanced distributions benefit smaller models \cite{11386287}, we apply per-category sampling with a 10,000-sample minimum. Specifically, each category contributes max(20\% × n, 10,000) samples, where n is the filtered category size. This prevents domination by high-volume categories and ensures diverse product coverage. The resulting dataset is split into training, validation and test sets in an 8:1:1 ratio, yielding approximately 2.1 million processed instances in total, including about 1.6 million training instances.

\subsection{Training Details}

In the training process, we use the BERT tokenizer with a maximum input length of 256 tokens. Each training instance contains one image, which is resized to 224x224 and normalized with ImageNet statistics. The proposed EffiMiniVLM model contains 27.7M active parameters and requires 6.8 GFLOPs, thereby maintaining a lightweight design. No additional acceleration or compression techniques are used and no external data beyond the competition-provided dataset are incorporated. The model is trained for up to 10 epochs with a batch size of 32, a dropout rate of 0.1 and a learning rate of $2 \times 10^{-5}$. A cosine learning-rate scheduler with a warmup ratio of 0.1 is employed, while gradient clipping with a maximum norm of 1.0 is applied for stability. To preserve label distribution during partitioning, stratification is performed according to the average rating, with rare rating values appearing 50 times or fewer grouped into a shared rare bucket. For optimization, we use a weighted Huber loss in which per-sample weights are derived from the logarithm of the rating number, normalized by the training-set mean and clipped at 5.0 to place greater emphasis on items with more ratings. The weighting scheme is formally defined as follows. For fair evaluation, the validation and test sets are assessed using the unweighted Huber loss. Model performance is measured using the Pearson Linear Correlation Coefficient (PLCC) and checkpoint selection is based on the best validation PLCC. Early stopping with a patience of one epoch is applied and most models converge after approximately four epochs.

Let $r_i$ be the rating number for sample $i$ in the training set.

The mean logarithmic rating number over all $N$ training samples is defined as:
\begin{equation}
\mu = \frac{1}{N}\sum_{j=1}^{N} \log(1 + r_j)
\end{equation}

Accordingly, the weight assigned to sample $i$ is computed as:
\begin{equation}
w_i = \min\left(\frac{\log(1 + r_i)}{\mu}, 5.0\right)
\end{equation}

Then the weighted training loss for a batch of size $B$ is given by:
\begin{equation}
L = \frac{1}{B}\sum_{i=1}^{B} w_i \cdot \mathrm{Huber}(\hat{y}_i, y_i)
\end{equation}

where $\hat{y}_i$ denotes the predicted rating and $y_i$ denotes the ground-truth average rating.

\section{Results and Discussions}

\subsection{Experimental Results}

To evaluate the effect of the proposed rating-count-based weighted Huber loss, we conduct a targeted comparison against the baseline setting using EfficientNet-B0 and MiniLM-L6-H384 on a 5\% subset of the training data, comprising approximately 420k samples without balanced main-category sampling. As shown in Table \ref{tab:weighted_huber}, introducing the weighted Huber loss improves the training PLCC from 0.3336 to 0.3649 and the validation PLCC from 0.3205 to 0.3306. This indicates that emphasizing samples with larger rating number provides a modest but consistent benefit.

\begin{table}[htbp]
\centering
\caption{Effect of weighted Huber loss on EffiMiniVLM}
\label{tab:weighted_huber}
\resizebox{\columnwidth}{!}{%
\begin{tabular}{lccc}
\hline
\textbf{Method} & \textbf{Model Size (M)} & \textbf{Train PLCC} & \textbf{Valid PLCC} \\
\hline
Baseline & 27.7 & 0.3336 & 0.3205 \\
Weighted Huber Loss & 27.7 & 0.3649 & 0.3306 \\
\hline
\end{tabular}%
}
\end{table}

To further investigate the impact of data scale and text encoder choice, Table \ref{tab:data_model_comparison} compares different configurations. The first three rows report results on the workshop development set for models trained with increasing amounts of data. The last two rows report workshop test-set results for the 1.6M-sample setting with different text encoders. The results show that performance improves consistently as the training data size increases, and replacing MiniLM-L6-H384 with MiniLMv2-L6-H384-distilled-from-BERT-Large further improves performance, achieving the best results of 0.36 PLCC and 0.40 CES.

\begin{table}[htbp]
\centering
\caption{Performance comparison of different data scales and text encoders on the workshop development and test sets}
\label{tab:data_model_comparison}
\scriptsize
\begin{tabular}{p{1.5cm} p{3.8cm} cc}
\hline
\textbf{Data Size} & \textbf{Model} & \textbf{PLCC} & \textbf{CES} \\
\midrule
\multicolumn{4}{c}{\textit{Workshop development set}} \\
\midrule
195k (1\%)  & MiniLM-L6-H384 & 0.30 & 0.33 \\
470k (5\%)  & MiniLM-L6-H384 & 0.32 & 0.36 \\
871k (10\%) & MiniLM-L6-H384 & 0.33 & 0.36 \\
\midrule
\multicolumn{4}{c}{\textit{Workshop test set}} \\
\midrule
1.6M (20\%) & MiniLM-L6-H384 & 0.35 & 0.39 \\
1.6M (20\%) & MiniLMv2-L6-H384-distilled-from-BERT-Large & 0.36 & 0.40 \\
\hline
\end{tabular}
\end{table}

\subsection{Efficiency Analysis}

To assess the runtime efficiency of the proposed solution, we measure inference performance on a Dell Pro Max 16 platform equipped with a single NVIDIA RTX PRO 2000 Blackwell GPU. Table \ref{tab:effiminivlm_efficiency} summarizes the complete set of efficiency metrics and experimental settings. All efficiency results are obtained using end-to-end inference timing with a batch size of 32 and five warm-up batches before measurement. The reported metrics are computed over 24,361 test samples, comprising a total of 4,840,498 input tokens.

EffiMiniVLM contains 27.7M total parameters, all of which are active during inference, with a computational cost of 6.8 GFLOPs. The model achieves a total runtime of 63.16 s, corresponding to an average inference latency of 2.59 ms/sample and 0.0130 ms/token. The measured throughput reaches 385.68 samples/s and 76,633.66 tokens/s. In addition, GPU memory usage remains low, with a peak allocated VRAM of 454.06 MB and a peak reserved VRAM of 678.00 MB. Overall, these results indicate that EffiMiniVLM delivers strong inference efficiency with low memory overhead, supporting the resource-efficient objective of the challenge.

\begin{table}[t]
\centering
\caption{Efficiency metrics of EffiMiniVLM}
\label{tab:effiminivlm_efficiency}
\scriptsize
\begin{tabular}{p{0.38\linewidth} p{0.52\linewidth}}
\toprule
\textbf{Metric} & \textbf{Value} \\
\midrule
Total Parameters (M) & 27.7 \\
Active Parameters (M) & 27.7 \\
FLOPs (G) & 6.8 \\
Hardware & Dell Pro Max 16 (NVIDIA RTX PRO 2000 Blackwell GPU) \\
Batch size & 32 \\
Timing scope & End-to-end \\
Warmup batches & 5 \\
Measured batches & 762 \\
Measured samples & 24,361 \\
Measured input tokens & 4,840,498 \\
Total runtime (s) & 63.16 \\
Latency (ms/sample) & 2.59 \\
Latency (ms/token) & 0.0130 \\
Throughput (samples/s) & 385.68 \\
Throughput (tokens/s) & 76,633.66 \\
Peak VRAM allocated (MB) & 454.06 \\
Peak VRAM reserved (MB) & 678.00 \\
Acceleration techniques & None \\
\bottomrule
\end{tabular}
\end{table}

\subsection{Quantitative and Qualitative Advantages}

\textbf{Quantitative Analysis}: Since our primary objective is to maximize resource efficiency, our method achieves the lowest resource cost on the leaderboard, at only 0.1. In comparison, teams ps3336 and skywalker5165 obtain a similar PLCC score of 0.37, which is only 0.01 higher than our score, while requiring approximately four times more computational resources. A similar pattern can also be observed in the higher performance regime. For example, team wleach achieves the same final score of CES 0.40 as our method, but with an eightfold increase in resource cost. In contrast, the top three teams on the leaderboard report resource costs ranging from 0.46 to 0.55, which is approximately around the median resource cost across all submissions. This highlights that competitive performance is typically associated with substantially higher computational budgets, whereas our method operates in a significantly lower resource regime.

\begin{figure}
    \centering
    \includegraphics[width=1\linewidth]{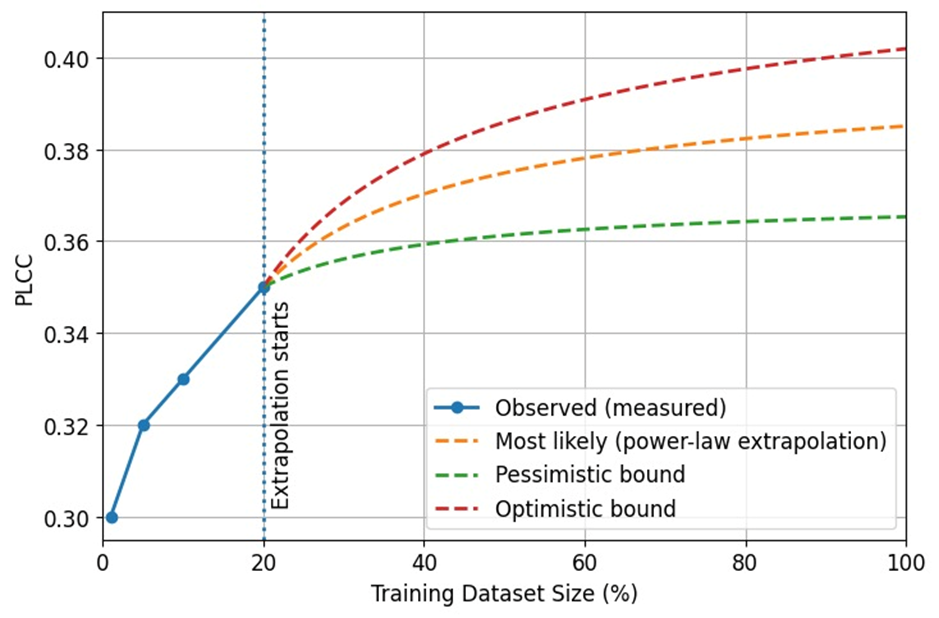}
    \caption{Extrapolation of the scaling behaviour with increasing training data}
    \label{fig:scaling_law}
\end{figure}

We further analyze the effect of training data scaling. As shown in Figure \ref{fig:scaling_law}, increasing the training data from 1\% to 20\% leads to a consistent improvement in PLCC, from 0.30 to 0.35. Empirical studies on neural scaling laws \cite{kaplan2020scalinglawsneurallanguage} suggest that model performance follows a sublinear power-law relationship with data size, with diminishing gains as model capacity becomes the limiting factor. Motivated by this behavior, we extrapolate the PLCC curve beyond the measured regime using a saturating power-law model. Extrapolation to 100\% data usage yields an estimated PLCC in the range of 0.365 to 0.402. When translated to the CES metric, this corresponds to a range of 0.4015 to 0.4422, with a most likely estimate of approximately 0.4235. This CES score of \~0.42 is sufficient to rival even the 3\textsuperscript{rd} spot on the leaderboard with the lowest resource cost of 0.1.

\begin{table*}[t]
\centering
\caption{Post-Challenge Exploration of the Impact of Dataset Scaling on a Custom Train/Validation Split}
\label{tab:dataset_scaling_custom_split}
\scriptsize
\begin{tabular}{lccccc}
\hline
Model & Data Used & Custom Val PLCC & Custom Val CES & Test PLCC & Test CES \\
\hline
MiniLM-L6-H384-uncased & 871K (10\%) & 0.3511 & 0.3862 & 0.33 & 0.36 \\
MiniLM-L6-H384-uncased & 1.6M (20\%) & 0.3635 & 0.3999 & 0.35 & 0.39 \\
MiniLMv2-L6-H384-distilled-from-BERT-Large & 1.6M (20\%) & 0.3648 & 0.4013 & 0.36 & 0.40 \\
MiniLMv2-L6-H384-distilled-from-BERT-Large & 3.3M (40\%) & 0.3884 & 0.4274 & -- & -- \\
\hline
\end{tabular}
\end{table*}

In fact, for exploratory purposes, we evaluated the impact of dataset scaling on a custom train/validation split based on the provided training set. As shown in Table \ref{tab:dataset_scaling_custom_split}, we observed that increasing the dataset utilization rate to 40\% improved model performance, achieving PLCC and CES scores of 0.3884 and 0.4272, respectively. These results demonstrate the viability of dataset scaling and align with our extrapolation results in Figure \ref{fig:scaling_law}, although they are based on a different evaluation set. Nevertheless, Table \ref{tab:dataset_scaling_custom_split} presents the results of scaling the dataset from 10\% to 40\% on a custom train/validation split, indicating the potential for further improvement with more data.

\textbf{Qualitative Analysis}: An important observation is that our approach is the only method achieving a final CES score of 0.40 or higher without using additional training datasets. This constraint is primarily due to limited computational resources, yet the results demonstrate that competitive performance can still be achieved under strict efficiency constraints. Moreover, our current implementation does not incorporate techniques such as knowledge distillation or the use of external datasets, which are adopted by several top-performing teams. This suggests that the current results do not yet reflect the full potential of the proposed framework. For instance, the extrapolation is based on EffiMiniVLM using MiniLM-L6-H384 as the text encoder. Replacing it with MiniLMv2-L6-H384-distilled-from-BERT-Large, as indicated in Table \ref{tab:data_model_comparison}, already provides measurable improvements. Also, Table \ref{tab:dataset_scaling_custom_split} demonstrates that our model can be further improved simply by scaling the dataset. Given its extremely low resource requirement, we believe that our EffiMiniVLM framework provides a strong foundation and could achieve even higher scores when combined with larger-scale training data.

\subsection{Novelty Degree}

To the best of our knowledge, the proposed method is the first to combine EfficientNet with MiniLM for a lightweight multimodal regression task. However, we observe similar models that use MiniLM for lightweight multimodal language modeling  \cite{wang2021minivlmsmallerfastervisionlanguage} and visual question answering (VQA)  \cite{noor2025lightweight}. The research work in \cite{wang2021minivlmsmallerfastervisionlanguage} uses EfficientDet for the visual branch, totaling 53.2M parameters. Meanwhile,  research work \cite{noor2025lightweight} uses a Vision Transformer (ViT) with 86M parameters as the vision encoder, coupling with MiniLM. In both cases, our proposed EffiMiniVLM with 27.7M parameters is significantly smaller than these two models, although the use cases differ.

\section{Conclusions}

In this work, we present EffiMiniVLM, a lightweight vision–language regression model designed with efficiency as the primary objective. Despite using only 27.7M parameters and 6.8 GFLOPs, our method achieves a final CES score of 0.40 while maintaining the lowest resource cost of 0.1 among all submissions. These results demonstrate that carefully designed compact multimodal architectures can remain competitive with significantly larger models. Our experiments further show that performance consistently improves as training data increases, suggesting that the current model has not yet reached its capacity limit. Extrapolation based on neural scaling behaviour indicates that EffiMiniVLM could achieve even higher performance when trained on the full dataset. Future work will explore scaling the training data, incorporating additional datasets and applying knowledge distillation to further improve performance while maintaining strong resource efficiency.

%%%%%%%%% REFERENCES
{\small
\bibliographystyle{ieeenat_fullname}
\bibliography{egbib}
}

\end{document}